\pdfoutput=1
\documentclass[letterpaper]{article} % DO NOT CHANGE THIS
\usepackage{aaai24}  % DO NOT CHANGE THIS
\usepackage{times}  % DO NOT CHANGE THIS
\usepackage{helvet}  % DO NOT CHANGE THIS
\usepackage{courier}  % DO NOT CHANGE THIS
\usepackage[hyphens]{url}  % DO NOT CHANGE THIS
\usepackage{graphicx} % DO NOT CHANGE THIS
\usepackage{natbib}  % DO NOT CHANGE THIS AND DO NOT ADD ANY OPTIONS TO IT
\usepackage{caption} % DO NOT CHANGE THIS AND DO NOT ADD ANY OPTIONS TO IT
\usepackage{algorithm}
\usepackage{algorithmic}
\usepackage{booktabs}
\usepackage{multirow}
\usepackage{makecell}
\usepackage{xcolor}
\definecolor{darkgreen}{RGB}{0,100,0}
\usepackage{array}

\usepackage{amsfonts}
\usepackage{amsmath}
\usepackage{newfloat}
\usepackage{listings}
\DeclareCaptionStyle{ruled}{labelfont=normalfont,labelsep=colon,strut=off} % DO NOT CHANGE THIS
\floatstyle{ruled}
\newfloat{listing}{tb}{lst}{}
\floatname{listing}{Listing}
\title{Turning Dust into Gold: Distilling Complex Reasoning Capabilities from LLMs by Leveraging Negative Data}
\author {
    Yiwei Li\textsuperscript{\rm 1}\footnotemark[1],
    Peiwen Yuan\textsuperscript{\rm 1}\footnotemark[1],
    Shaoxiong Feng\textsuperscript{\rm 2},
    Boyuan Pan\textsuperscript{\rm 2},
    Bin Sun\textsuperscript{\rm 1},
    Xinglin Wang\textsuperscript{\rm 1},
    Heda Wang\textsuperscript{\rm 2},
    Kan Li\textsuperscript{\rm 1}\footnotemark[2]
}

\affiliations {
    \textsuperscript{\rm 1} School of Computer Science, Beijing Institute of Technology \quad
    \textsuperscript{\rm 2} Xiaohongshu Inc \\
    \{liyiwei,peiwenyuan,binsun,wangxinglin,likan\}@bit.edu.cn \\\{shaoxiongfeng2023,whd.thu\}@gmail.com \{panby\}@zju.edu.cn
}

\begin{document}

\maketitle
\renewcommand{\thefootnote}{\fnsymbol{footnote}} 
\footnotetext[1]{Equal contributions.} 
\footnotetext[2]{Corresponding author.} 

\renewcommand{\thefootnote}{\arabic{footnote}}
\begin{abstract}
Large Language Models (LLMs) have performed well on various reasoning tasks, but their inaccessibility and numerous parameters hinder wide application in practice. 
One promising way is distilling the reasoning ability from LLMs to small models by the generated chain-of-thought reasoning paths. 
In some cases, however, LLMs may produce incorrect reasoning chains, especially when facing complex mathematical problems. 
Previous studies only transfer knowledge from positive samples and drop the synthesized data with wrong answers. 
In this work, we illustrate the merit of negative data and propose a model specialization framework to distill LLMs with negative samples besides positive ones. 
The framework consists of three progressive steps, covering from training to inference stages, to absorb knowledge from negative data.
We conduct extensive experiments across arithmetic reasoning tasks to demonstrate the role of negative data in distillation from LLM\footnotemark[1]. 
\end{abstract}

\footnotetext[1]{Our code and data have been released on \url{https://github.com/Yiwei98/TDG}.}
\section{Introduction}
Nowadays, owing to chain-of-thought (CoT) prompting \citep{COT}, large language models (LLMs) exhibit strong reasoning capabilities \citep{AGI}, especially when it comes to complex mathematical problems \citep{MATH}.
Unfortunately, CoT has been demonstrated to be an emergent property of models with more than 100B parameters, but not of smaller models \citep{Emergent}. The burdensome computational requirements and high inference costs of these models hinder their development in real-world scenarios with limited resources \citep{ReasonTeacher}. Thus, the goal of our research is to enable complex arithmetic reasoning in small models for deploying at scale.

\begin{table}[th]
    \centering
    \small
    \begin{tabular}{l c c c c}
    \toprule
    MATH Dataset &  Intersection & Pos & Neg &IoU\\ \midrule
    InterAlgebra &  4  & 35  &  21 & 0.077\\ 
    Prealgebra &  9  & 72  & 43  & 0.085\\
    Geometry &  1  & 21  &  10 & 0.033\\
    NumberTheory &  1  & 29  & 17  &0.022\\
    Precalculus &  2  & 25  &  16 &0.051\\
    Probability &  4  & 19  &  16 & 0.129\\
    Algebra &  8  & 52  &  43 & 0.062\\
    Overall &  29  & 253  &  166 & 0.074\\
        \bottomrule
    \end{tabular}
    \caption{The distribution of correct answers in MATH test set. Pos and Neg refer to models trained on positive and negative samples respectively. Intersection over Union (IoU) exhibits a remarkably low value across all subsets, which confirms the value of negative samples.}
    \label{tb:intro}
\end{table}

Knowledge distillation \citep{KD} offers a promising way to transfer specific capabilities from LLMs into smaller models. 
This process is also referred to as model specialization enforcing compact models to focus on certain skills. 
Prior works \citep{TeachReason,Specialize,DistillStep} employed LLMs with in-context learning (ICL) \citep{GPT-3} to generate reasoning paths (rationales) of math problems, which are more beneficial for small models to acquire complex reasoning ability than reference reasoning paths. 
Table~\ref{tb:intro} shows an intriguing phenomenon: models trained on positive and negative data separately have an extremely small overlap (intersection) in their correct answers on the MATH test set. 
Although the negative model has a lower accuracy, it can address some questions that the positive model is unable to provide correct answers, which confirms the valuable knowledge contained in negative data.
Additionally, the undesirable behaviors within negative data are also useful when preventing the model from committing similar issues.
Another reason that we should exploit negative data is the token-based pricing strategy of OpenAI.
Even for GPT-4, the accuracy on MATH dataset is less than 50\% \citep{GPTresult}, meaning that all tokens of negative data are charged for nothing.
Therefore, instead of discarding negative samples, we extract and utilize valuable knowledge from negative samples to boost the model specialization. 

The conventional process of model specialization can be summarized as three steps \citep{PaD}: 
The first step is chain-of-thought distillation, training small models with reasoning chains generated from LLMs. 
The second step can be regarded as self-enhancement, conducting self-distillation \citep{SelfDistill} or self-augmentation to further optimize the models. 
Besides, self-consistency \citep{SC} is widely used as an effective decoding strategy to boost the model performance in reasoning tasks.
In this work, we propose a novel model specialization framework (shown in Figure~\ref{pic:overview}) that can exploit negative data to enhance the distillation of the complex reasoning abilities from LLMs.
Specifically, we first develop the negative assistant training (NAT) approach, where dual LoRA \citep{LORA} structure is designed to capture knowledge from both positive and negative sides. As an auxiliary module, the knowledge of negative LoRA can be dynamically integrated into the training process of positive LoRA through a corrected attention mechanism.
For self-enhancement, we devise negative calibrated enhancement (NCE), which regards the negative output as a baseline to strengthen the distillation of critical positive rationales.
In addition to the training stage, we also leverage the negative information during inference.
Traditional self-consistency allocates equal or probability-based weights to all candidate outputs, leading to some fallible answers being voted up. To alleviate this issue, adaptive self-consistency (ASC) is proposed to conduct ranking before voting, where the ranking model is trained on both positive and negative data. 

We perform comprehensive experiments and detailed analyses across arithmetic reasoning tasks with LLaMA-7b \citep{LLaMA} as the student model. 
Previous model specialization work only validated on ordinary datasets (e.g., GSM8K, ASDiv, etc.), while we are the first to focus on the challenging competition mathematical problems -- MATH dataset \citep{MATH}. 
Experiments show that: 
(1) Negative assistant training can provide a more comprehensive way to absorb the knowledge from negative data.
(2) Negative calibrated enhancement can make the process of self-distillation more targeted on crucial knowledge. 
(3) Ranking model trained on both positive and negative rationales can assign appropriate weights for answer aggregation.
In summary, key contributions of this work are as follows:
\begin{itemize}
    \item We illustrate that negative samples with incorrect answers can also provide a valuable resource besides positive data for distilling knowledge from LLMs in complex arithmetic reasoning tasks.
    \item To fully leverage the negative data, we propose a model specialization framework consisting of three progressive steps, spanning from training to inference stages. 
    \item Extensive evaluations on challenging arithmetic reasoning dataset demonstrate that the proposed framework can effectively exploit the negative information and outperform baselines by a large margin.
\end{itemize}

\section{Related Work}
\subsection{Chain-of-Thought Reasoning}
The approach of solving complex reasoning problems by generating chain-of-thought (rationales) has been proven to be an effective method \citep{COT}. By following the pattern of gradually solving sub-problems, both few-shot CoT \citep{Specialize} and zero-shot CoT \citep{ZeroCOT} can stimulate the potential reasoning ability of LLMs. On this basis, Least-to-most prompting \citep{LeastToMost} suggests explicitly splitting the problem and solving them step by step. Self-Consistency \citep{SC} further improves accuracy by conducting vote between multiple diverse rationales. PHP proposes \citep{PHP} iteratively generating answers and adding the historically generated answers as hints to the context to achieve the final convergence on the answer. Both correct and incorrect answers generated during this iteration process serve as hints to provide effective information. We also think that responses with incorrect answers from LLMs can provide valuable information, but the differences lie in: (1) We believe that not only the generated answers, but also the rationales contain valuable knowledge. (2) We consider utilizing these negative samples in the process of transferring knowledge from LLMs to smaller models instead of only inference stage.

\subsection{Knowledge Distillation from Large Model}
Knowledge distillation \citep{KD,DisBERT} has proven effective for transferring knowledge from a large model to a smaller one. This process is usually achieved by optimizing the parameters of smaller models so that their outputs (distributions \citep{BD}, hidden states \citep{TinyBERT}, attentions \citep{DisfromBERT}) can be closer to that of large models. However, the black-box nature of current mainstream LLMs (e.g., GPT4) hinders the application of these methods. Thus, many studies \citep{ReasonTeacher,Specialize,PaD} have attempted to conduct hard distillation by fine-tuning smaller models directly on the LLMs generated responses with correct answers. However, as previously mentioned, responses generated by LLMs that contain incorrect answers also contain valuable knowledge. Discarding this portion of data directly would be a pity, especially considering that a significant portion of responses generated by LLMs in complex reasoning tasks end with incorrect answers. To this end, we propose multiple methods to fully utilize these \textit{abandoned knowledge} in the process of transferring reasoning abilities of LLMs.

\subsection{Learning From Negative Views}
Samples that reflect some particular undesirable behavior are called negative data, which has been studied to help model correct such behavior \citep{NT,UL,IUL}. \citet{NT} conducts negative updates with training signals provided by negative samples to avoid model generating such data. \citet{UL,ULDia} penalizes the model for outputting words with certain characteristics by introducing an unlikelihood loss term.
\citet{ND,CDL} suggests maximizing the distance between the predictions of the negative teacher and student. 
These methods only consider the use of negative training signals in negative samples. But in fact, negative data can also provide valuable positive knowledge. In this work, we investigated how to comprehensively utilize knowledge of negative data from both positive and negative perspectives.

\section{Methodology}

\begin{figure*}[t]
\centering
\includegraphics[width=0.90\textwidth]{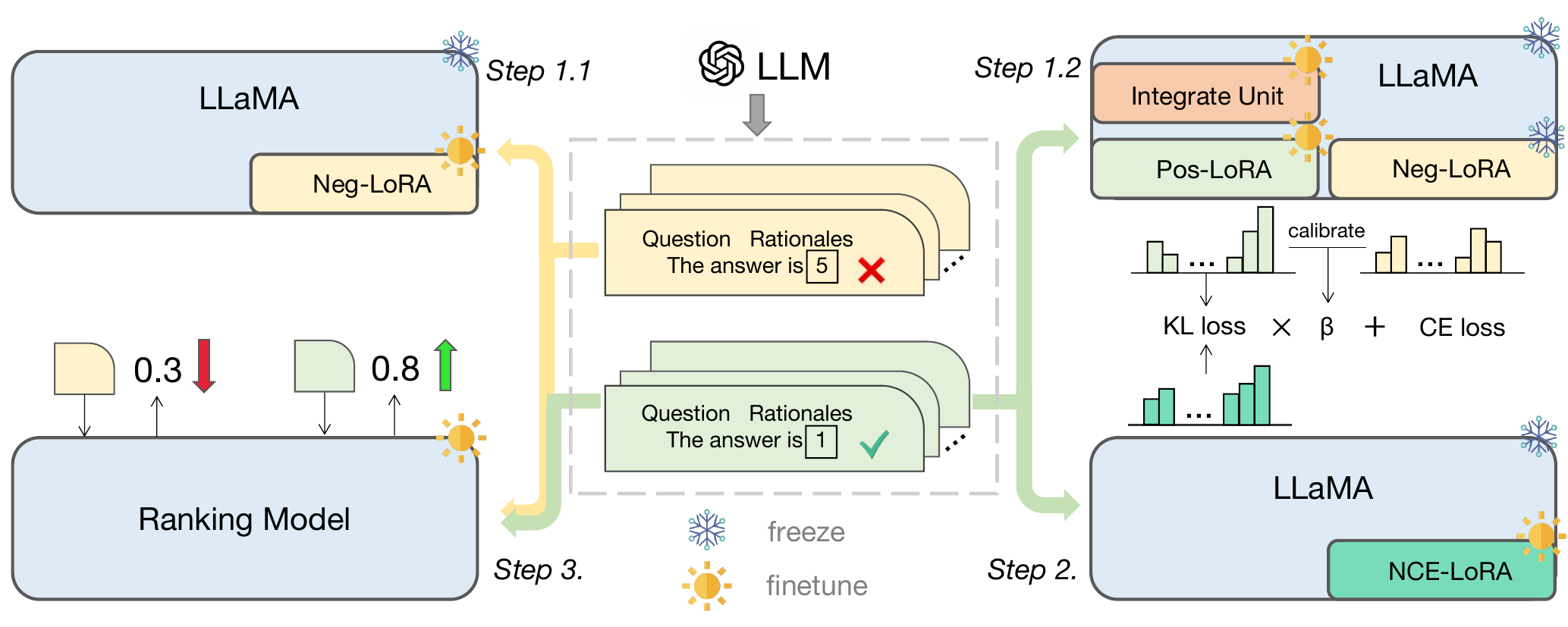} % Reduce the figure size so that it is slightly narrower than the column.
\caption{\textbf{The overview of proposed framework. Step 1}: Training Neg-LoRA on negative samples to assist in the learning of reasoning on positive data through Integrate Unit. \textbf{Step 2}: Utilizing Neg-LoRA as baseline to calibrate the process of self-enhancement. \textbf{Step 3}: Training a ranking model on both positive and negative samples. Then weighting the candidates adaptively during inference according to scores from it.
}
\label{pic:overview}
\end{figure*}

\subsection{Background}

\subsubsection{Chain-of-Thought Distillation}
\label{sec:cot-dis}
\citet{Explain} demonstrated that training a language model on a dataset with explicit rationales preceding the answer could improve the ability to generate the final answer. 
Thus, chain-of-thought distillation is proposed to maximize the manifestation of the reasoning abilities of the LLMs on smaller models.
Denote $\mathcal{D} = \{(x_i, y_i)\}^N$ to be a dataset with $N$ training instances, where $x_i$ is a problem and $y_i$ is its answer. Given an additional set of M demonstrations $\{d_i =(x_i^d,r_i^d,y_i^d)\}^M$, where $r$ represents rationales, the prompt $\{p_i=(d_1,...,d_M,x_i)\}^N$ is input to the LLM and obtain responses $\{(\hat{r}_i,\hat{y}_i)\}^N$. Previous work \citep{ReasonTeacher, Specialize, STaR} retain positive samples $S_{pos}$ where $\hat{y}=y$, and maximizes the likelihood of the student model to generate $(\hat{r},\hat{y})$ as follows:

\begin{equation}
\mathbb{E}_{(x,\hat{r},\hat{y}) \sim S_{pos}}  \mathrm{log}  P (\hat{y},\hat{r} | x;\theta).
\label{eq:likelihood}
\end{equation}

\subsubsection{Self-Enhancement}
Based on the idea of human self reflection to achieve progress, various methods \citep{selfimprove,selfaug,SelfDistill} have been proposed to strengthen language models based on their own knowledge, which we collectively refer to as self-enhancement. It consists two common methods: one is self-augmentation \citep{selfaug}, where the model first generates data with diversity and then trains on them to achieve better generalization \citep{selfimprove}. 
The other is self-distillation \citep{PaD}, which involves using the model itself as teacher to complete iterative distillation, thereby utilizing dark knowledge to further improve the performance.

\subsubsection{Self-Consistency}

Self-consistency \citep{SC} capitalizes on the notion that a intricate problem requiring logical thinking usually offers several distinct approaches that all lead to the same accurate answer. Based on this, 
multiple candidates $\{(\hat{r}^l,\hat{y}^l)\}^L$ to problem $x$ are suggested to generate through sampling, and the most consistent $\hat{y}$ is selected as the final prediction through a voting process:

\begin{equation}
\hat{y} = \arg\max_{i} \sum_{l=1}^{L} \mathbb{I}(\hat{y}^l = i)
\label{eq:sc}
\end{equation}
where $\mathbb{I}(\hat{y}^l = i)$ is the indicator function (equal to 1 when $\hat{y}^l$ is equal to answer $i$, and 0 otherwise).

\subsection{Negative Assistant Training}
As shown in Table~\ref{tb:intro}, negative samples also contains valuable knowledge, which can even serve as a good complement to positive data. However, there is an increased risk of inference errors for $\hat{r}$ corresponding to negative data where $\hat{y} \neq y$. Extracting useful knowledge from negative samples without being affected by undesirable behaviors is therefore a challenging task. To address this, we propose a two-stage Negative Assistant Training (NAT) Paradigm (\textit{Step 1.1 and 1.2} in Figure~\ref{pic:overview}).

\subsubsection{Absorbing Negative Knowledge}

First, we acquire the $(x,\hat{r},\hat{y})$ triplets from the LLM on mathematical problems as described in Background. To facilitate comprehensive learning of diverse problem-solving approaches from the LLM, we collect 8 distinct responses for each question, and categorizing these samples into $\mathcal{D}_{pos}$ and $\mathcal{D}_{neg}$ based on whether $\hat{y}$ equals to $\hat{y}$. 
Directly finetuning LLaMA on the union of $\mathcal{D}_{pos}$ and $\mathcal{D}_{neg}$ will inevitably introduce undesirable behaviors into the model. Therefore, we consider training a negative model on $\mathcal{D}_{neg}$ first, and extracting useful knowledge from it afterwards.
We choose LoRA module \citep{LORA} for its parameter efficient characteristics to finetune LLaMA on $\mathcal{D}_{neg}$ by maximizing the following expectation:

\begin{equation}
\mathbb{E}_{(x,\hat{r},\hat{y}) \sim \mathcal{D}_{neg}}  \mathrm{log}  P (\hat{y},\hat{r} | x;\theta_{neg}).
\label{eq:neglora}
\end{equation}
During this process, the parameters of LLaMA remain frozen, while the knowledge of $\mathcal{D}_{neg}$ is absorbed by LoRA $\theta_{neg}$. We denote LLaMA with $\theta_{neg}$ as $\mathcal{M}_{neg}$. 

\subsubsection{Dynamic Integration Unit}
\begin{figure}[t]
\centering
\includegraphics[width=0.4\textwidth]{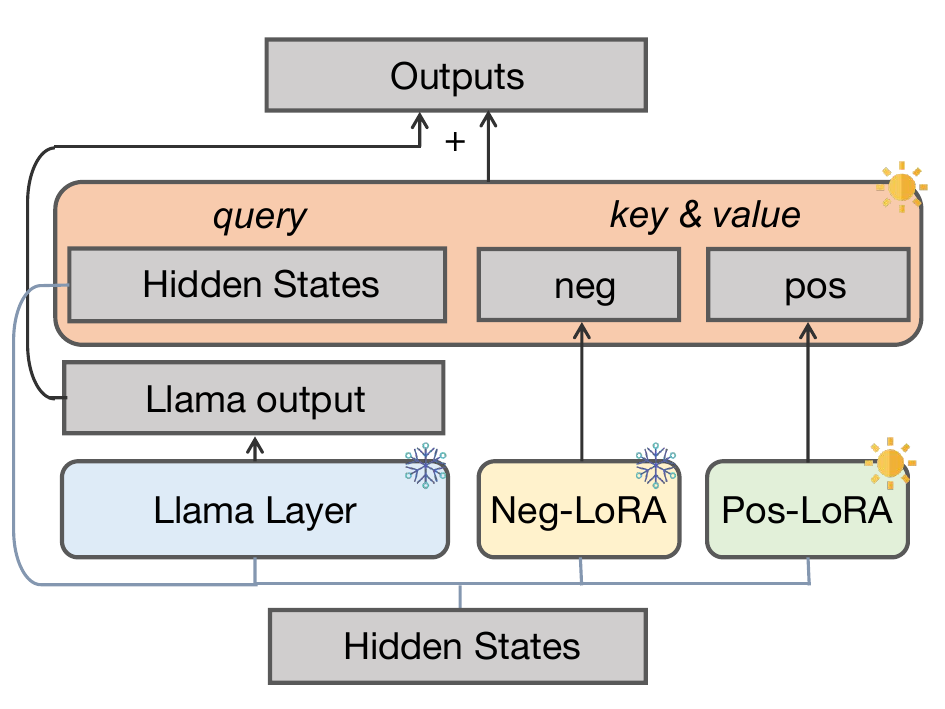} % Reduce the figure size so that it is slightly narrower than the column.
\caption{The workflow of Integrate Unit. The outputs of both Neg-LoRA and Pos-LoRA are fused through a corrected attention mechanism.}
\label{pic:integrate}
\end{figure}

Since it is impossible to pre-determine which mathematical problems $\theta_{neg}$ excels at, we design Dynamic Integrate Unit as shown in Figure~\ref{pic:integrate} to dynamically integrate knowledge from $\theta_{neg}$ during the learning process of positive knowledge in $\mathcal{D}_{pos}$. 
We freeze $\theta_{neg}$ to prevent the knowledge inside from being forgotten and additionally introduce positive LoRA modules $\theta_{pos}$. In each layer of LLaMA, we denote the output values obtained from $\theta_{pos}$ ($\theta_{neg}$) as $h_{pos}$ ($h_{neg}$) for input hidden states $h_{input} \in \mathbb{R}^{d}$. Ideally, we should positively integrate $h_{pos}$ and $h_{neg}$ to complement the beneficial knowledge in $\mathcal{D}_{neg}$ with respect to $\mathcal{D}_{pos}$ if $h_{neg}$ contains beneficial knowledge. When $h_{neg}$ contains detrimental knowledge, we should negatively integrate $h_{pos}$ and $h_{neg}$ to assist in reducing the possible undesirable behaviors within $\mathcal{D}_{pos}$. 
We propose a corrected attention mechanism to achieve this vision as follows: 

\begin{equation}
\alpha = W_Q(h_{input}) W_K([h_{pos};h_{neg}])^T+[0.5;-0.5]
\label{eq:attention1}
\end{equation}

% \begin{equation}
% h_{output} = \alpha \cdot W_V^1(W_V^2([h_{pos};h_{neg}]))
% \label{eq:attention2}
% \end{equation}
\begin{equation}
h_{output} = \alpha \cdot W_V([h_{pos};h_{neg}]))
\label{eq:attention2}
\end{equation}
where $W_Q,W_K\in \mathbb{R}^{d \times w}$ and $W_V \in \mathbb{R}^{d \times d}$ are trainable parameters during this stage 
(we use a bottleneck structure for $W_V$ for reducing parameters).
In Eq.~\eqref{eq:attention1}, we use $h_{input}$ as the query to calculate attention weights for $h_{pos}$ and $h_{neg}$. By adding a correction term [0.5;-0.5] on [$\alpha_{pos};\alpha_{neg}$], the attention weights for $h_{neg}$ are constrained to the range of [-0.5, 0.5], thus achieving the effect of adaptively integrating knowledge from $h_{neg}$ in both positive and negative directions. Ultimately, the sum of $h_{output}$ and the LLaMA layer outputs forms the output of the Dynamic Integrate Unit.

By employing NAT, $\mathcal{M}_{NAT}$ can inherit LLM's knowledge more comprehensively in both dimensions of diversity (more samples) and type (both positive and negative data), leading to improved complex reasoning abilities.

\subsection{Negative Calibrated Enhancement}
To further strengthen the reasoning ability of the model, we propose Negative Calibrated Enhancement (NCE) that use negative knowledge to aid with the self-enhancement process \citep{PaD,selfimprove}.
We first use $\mathcal{M}_{NAT}$ to generate $n$ ($\hat{r}, \hat{y}$) pairs for each problem in $\mathcal{D}$ as augmentation samples and supplement them to $\mathcal{D}_{pos}$. 
As for self-distillation \citep{PaD,FastBert}, we notice that some samples may contain more critical reasoning steps that can distinguish reasoners with different abilities. 
Our primary objective is to identify these key rationales and enhance the learning of them during self-distillation.
Considering that $\mathcal{M}_{NAT}$ has incorporated the valuable knowledge from $\mathcal{M}_{neg}$, the key insights that make $\mathcal{M}_{NAT}$ a superior reasoner than $\mathcal{M}_{neg}$ are implicit in the inconsistent rationale generating distributions between the two. Therefore, we use KL divergence to measure such inconsistency and maximize the expectation in Eq.~\eqref{eq:nce}, where $\theta_{NCE}$ denotes the LoRA module of the student model $\mathcal{M}_{NCE}$.
\begin{equation}
f_{\text{KL}}(\theta_{1},\theta_{2}) = \mathrm{KL}(P(\hat{r},\hat{y}|x;\theta_{1}),P(\hat{r},\hat{y}|x;\theta_{2}))
\label{eq:kl}
\end{equation}
\begin{equation}
\beta = \mathrm{Tanh}(f_{\text{KL}}(\theta_{neg},\theta_{NAT}))
\label{eq:beta}
\end{equation}
\begin{equation}
%\mathcal{L}_{NCE} = \beta * \mathrm{KL}(P(r|x;\theta_{NAT}),P(r|x;\theta_{NCE})) + \mathrm{CE} 
\mathbb{E}_{(x,\hat{r},\hat{y}) \sim S_{pos}}  \mathrm{log}  P (\hat{y},\hat{r} | x;\theta_{NCE}) + \beta * f_{\text{KL}}(\theta_{NAT},\theta_{NCE})
\label{eq:nce}
\end{equation}
A larger $\beta$ value indicates a greater divergence between $\mathcal{M}_{NAT}$ and $\mathcal{M}_{neg}$, implying that the generating distribution of $\mathcal{M}_{NAT}$ contains more crucial knowledge.
By introducing $\beta$ to adjust the loss weights of different samples, $\mathcal{M}_{NCE}$ will be able to learn selectively and generalize the knowledge embedded in $\mathcal{M}_{NAT}$.

\subsection{Adaptive Self-Consistency}
Self-Consistency (SC) technique \citep{SC} is effective for further improving the performance of models in complex reasoning \citep{PaD}. However, current methods either naively assign equal weights to each candidate or simply assign weights based on generation probabilities. 
These strategies fail to adjust the weights of candidates based on the quality of ($\hat{r},\hat{y}$) during the voting phase, which could potentially obscure the correct candidates.
To this end, we propose Adaptive Self-Consistency (ASC), which utilizes $\mathcal{D}_{neg}$ and $\mathcal{D}_{pos}$ to train a Ranking Model $\mathcal{M}_{rank}$ that can adaptively reweight candidates with justification.

\begin{table*}[t]
    \centering
    \small
    \begin{tabular}{c c c c c c c c c | c}
    \toprule
    Models & Methods & \makecell[c]{Counting\&\\Probability} &  \makecell[c]{Inter\\Algebra} & \makecell[c]{Number\\Theory} & Precalculus & Prealgebra & Geometry & Algebra & Average 
    \\ \midrule
    PaLM 62B & Few-shot  & - & - & - & - & - & - & - & 4.4 \\
    PaLM 540B & Few-shot  & - & - & - & - & - & - & - & 8.8 \\
    GPT3 175B & Few-shot  & 4.7 & 4.4 & 4.4 & 4.0 & 7.7 & 3.1 & 6.0 & 5.2\\
    GPT3 13B & Fine-tune & 4.1 & 4.7 & 5.5 & 5.8 & 6.8 & 7.1 & 5.3 & 5.6\\ \midrule
    LLaMA 7B& Fine-tune & 2.96 & 3.58 & 2.96 & 3.85 & 4.61 & 3.46 & 4.56 & 3.88 +0\%\\ \midrule
   \multirow{7}{*}{\makecell[c]{GPT-3.5\\Turbo\\CoT}} & CoT KD & 4.15 & 4.17 & 5.37 & 4.58 & 8.82 & 4.54 & 4.61 & 5.29 +36.3\%\\
    &MIX & 3.49 & 1.43 & 1.67 & 1.46 & 5.27 & 2.59 & 4.05 & 3.03 -21.9\%\\
    &CL & 4.64 & 3.93 & 5.74 & 4.03 & 7.39 & 2.51 & 5.98 & 5.16 +33.0\%\\
    &NT & 3.93 & 3.93 & 6.30 & 2.20 & 6.69 & 4.10 & 5.17 & 4.48 +15.4\%\\
    &UL & 4.98 & 3.86 & 5.37 & 3.85 & 6.70 & 4.10 & 5.27 & 4.96 +27.8\%\\
    &NAT & 5.70 & 5.24 &  6.67 & 3.85 & 9.99 & 5.64 & 7.94 & 6.81 +75.5\% \\ \midrule
    \multirow{7}{*}{\makecell[c]{GPT-4\\CoT}} &CoT KD & 3.71 & 4.88 & 6.30 & 3.30 & 6.56 & 3.67 & 7.73 & 5.59 +44.1\%\\
    &MIX & 3.28 & 2.86 & 2.96 & 4.21 & 5.45 & 3.55 & 6.66 & 4.49 +15.7\%\\
    &CL & 4.15 & 3.67 & 5.00 & 3.11 & 7.90 & 5.43 & 5.98 & 5.24 +35.1\%\\
    &NT & 3.28 & 2.46 & 4.07 & 3.85 & 8.92 & 6.05 & 5.97 & 5.14 +32.4\%\\
    &UL & 4.15 & 3.46 & 6.67 & 3.11 & 8.67 & 5.18 & 8.25 & 6.03 +55.4\%\\
    &NAT & 6.11 & 4.65 & 5.56 & 4.58 & 8.50 & 4.92 & 9.78 & 6.83 +76.0\% \\ 
     \bottomrule
    \end{tabular}
    \caption{Experimental results (\%) on MATH test set for NAT. We report the accuracy (solving rate) of math problems for each test set. Average is the mean value of all subjects. GPT3 and PaLM are from \citet{MATH} and \citet{SQR}, respectively. Comparing with standard fine-tune, NAT achieves about 75.75\% increase.}
    \label{tb:main_exp}
\end{table*}

\subsubsection{Ranking Model Training.}
Ideally, we hope that $\mathcal{M}_{rank}$ assigns higher weights to rationales that lead to the correct answer and vice versa.
Thus, we construct training samples $\{(p_i,q_i)\}^N$ in the following way:
\begin{equation}
(p,q)=
\begin{cases}
([x,\hat{y},\hat{r}],1) \ \ if \ \  \hat{y}=y
 \\
([x,\hat{y},\hat{r}],0) \ \ if \ \  \hat{y}\neq y
\end{cases}
\label{eq:rmdata}
\end{equation}
and use MSE loss to train $\mathcal{M}_{rank}$:
\begin{equation}
\mathcal{L}_{RM} = \sum_{i=1}^{N}\|\mathcal{M}_{rank}(p_i)-q_i\|_{2}
\label{eq:rm}
\end{equation}
\subsubsection{Weighting Policy.}
Building upon the foundation of $\mathcal{M}_{rank}$, we revise Eq.~\eqref{eq:sc} to Eq.~\eqref{eq:asc} to achieve the vision of adaptively reweighting the candidates reasonably.

\begin{equation}
\hat{y} = \arg\max_{i} \sum_{l=1}^{L}  \mathbb{I}(\hat{y}^l = i) \cdot \mathcal{M}_{rank}([x,\hat{y}^l,\hat{r}^l]) 
\label{eq:asc}
\end{equation}
From the view of knowledge transfer, ASC achieves further utilization of knowledge ( positive and negative) embedded in LLMs to help smaller models attain better performance.

\section{Experiments}

\subsection{Experimental Setup}

This work focuses on the challenging competition mathematical dataset MATH \citep{MATH}, which has 12,500 problems in total spanning seven various subjects.
Besides, the following four datasets are introduced to evaluate the generalization ability on out-of-distribution (OOD) data of the proposed framework: GSM8K \citep{GSM8K}, ASDiv \citep{asdiv}, MultiArith \citep{multi}, and SVAMP \citep{svamp}.
Detailed data statistics are shown in Appendix.

For teacher model, we use gpt-3.5-turbo and gpt-4 API from OpeanAI to synthesize reasoning chains. 
Given that the problems of MATH are challenging, we select LLaMA-7b as the student model.

There are two main types of baselines in our study: one includes LLMs, while the other is based on LLaMA-7b. In the case of LLMs, we compare with two popular models: GPT3 \citep{GPT-3} and PaLM \citep{Palm}.
As for LLaMA-7b, we first provide a comparison of our method with three settings: Few-shot, Fine-tune (on original training samples), CoT KD (chain-of-thought distillation). In terms of learning from negative views, four baseline methods will be further included: MIX (directly trains LLaMA with the mixture of both positive and negative data), CL (contrastive learning), NT (negative training) \citep{NT} and UL (unlikelihood) \citep{UL}. Please see Appendix for the details of these baselines. The evaluations of NCE and ASC will also include some other baselines that will be introduced in corresponding parts.

\subsection{Main Results}

\subsubsection{Native Assistant Training}
The evaluation results of NAT are presented in Table~\ref{tb:main_exp}, with all methods using greedy search (i.e. temperature = 0). It shows that proposed method NAT improves task accuracy across all baselines. It can be seen from the low values of GPT3 and PaLM that MATH is a very difficult math dataset, but NAT can still accomplish competitive performance with much less parameters. Comparing with fine-tuning on the original data, NAT achieves about 75.75\% increase under two different CoT sources. In comparison with CoT KD on positive samples, the mainstream specialization pattern, NAT also improves accuracy significantly, demonstrating the value of negative samples. As for baselines to utilize negative information, the lowest performance of MIX suggests that directly training the negative samples will make model toxic. Other methods are also mostly inferior to NAT, which indicates that using negative samples only in the negative direction is not sufficient in complex reasoning tasks.

\begin{figure}[t]
\centering
\includegraphics[width=0.45\textwidth]{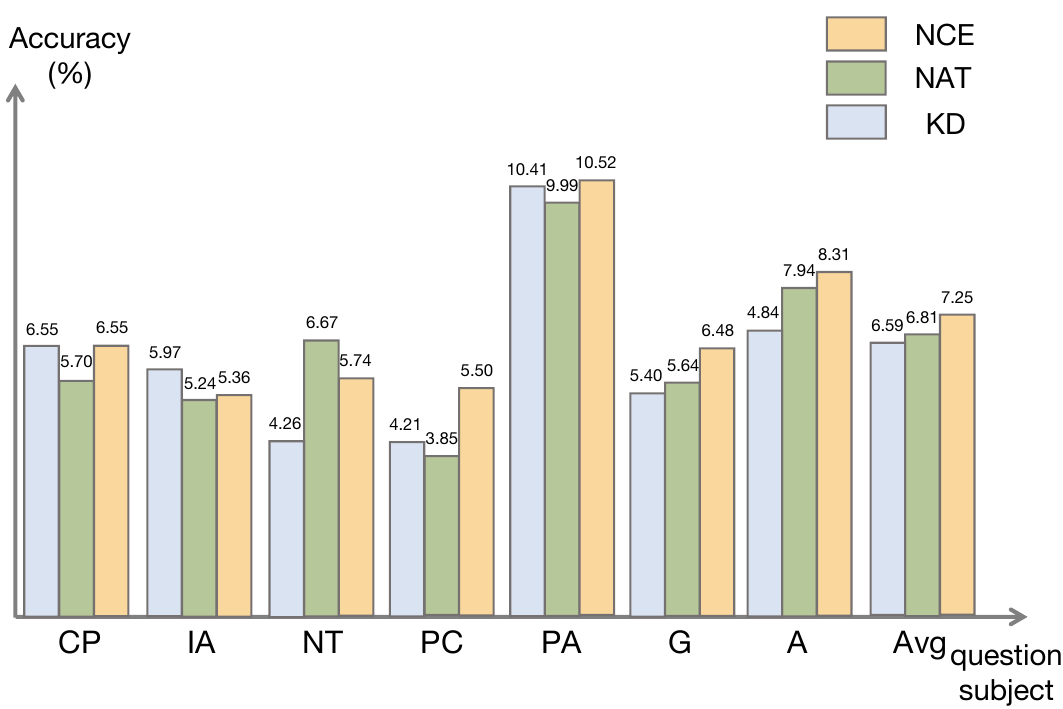} % Reduce the figure size so that it is slightly narrower than the column.
\caption{Experimental results (\%) of NCE. KD denotes knowledge distillation with data augmentation.
}
\label{pic:kd}
\end{figure}

\subsubsection{Negative Calibrated Enhancement}

The main results under the data of gpt-3.5-turbo of NCE are shown in Figure~\ref{pic:kd}. 
Compared with knowledge distillation (KD), NCE achieves an average progress of 10\% (0.66), which demonstrates the effectiveness of distillation with calibration information offered by negative samples. Although NCE reduced some parameters (e.g., Neg-LoRA) compared to NAT, it still achieved a progress of 6.5\% (0.44), implementing compressed model and improved performance.

\subsubsection{Adaptive Self-Consistency}

\begin{table}[t]
    \centering
    \small
        \begin{tabular}{c c c c c c c c}
    \toprule
    Models & Strategies & CP & NT & PC & A* &Ave\\ \midrule
    \multirow{3}{*}{CoT KD}& SC & 7.38  & 6.62 & 5.70 & 8.70 & 7.85 \\
     & SC \textit{w}WS & 7.22  & 6.64 & 5.75 & 8.52 & 7.82 \\
    & ASC & 7.70 & 6.97 & 6.16  & 9.12 & 8.25\\ \midrule
    \multirow{3}{*}{NAT} & SC & 8.65 & 7.49 & 5.75  & 11.14 & 9.25 \\
    & SC \textit{w}WS  & 8.67 & 7.34 & 5.77  & 11.08 & 9.21 \\
    & ASC & 9.30 & 8.33 & 5.83 & 11.88 & 9.84 \\ \midrule
    \multirow{3}{*}{NCE} & SC & 9.21  & 7.94 &  5.96 & 11.32 & 9.69 \\
    & SC \textit{w}WS & 9.13 &7.84 & 5.99& 11.25 & 9.64 \\
    & ASC & 9.87 & 8.21 & 6.37 & 11.89 & 10.23\\ 
    \bottomrule
    \end{tabular}
    \caption{Experimental results (\%) on MATH for ASC. A* is the average of InterAlgebra, Prealgebra and Algebra.}
    \label{tb:ASC}
\end{table}

To evaluate ASC, we compare it with base SC and its weighted sum (WS) version. 
We generate 16 samples with sampling temperature $T=1$. The results from Table~\ref{tb:ASC} shows that ASC is a more promising strategy to aggregate the answers from different samples. SC with WS doesn't outperform base SC, which is consistent with \citet{SC}.
Note that the accuracy of ranking model is only about 60\%, indicating that the performance of ASC can be further improved with higher accuracy. Refer to Accuracy of Ranking Model for detailed analysis.

\subsection{Analysis}
In order to better understand the usefulness of the negative knowledge and the effectiveness of our framework, we carry out extensive analysis on LLaMA distilled from gpt-3.5-turbo in terms of both quantitative and qualitative measures.

\subsubsection{Generalization}
\begin{table}[th]
    \centering
    \small
        \begin{tabular}{l c c c c}
    \toprule
    Methods & GSM8K & ASDiv & MultiArith & SVAMP \\ \midrule
    Fine-tune & 17.51 & 36.37 &53.17 &17.90 \\
    CoT KD & 38.81 & 76.43 & 83.5 & 47.40  \\
    NAT & 41.24 & 76.11 & 84.67 & 47.20  \\
    KD & 41.55 & 75.86 & 88.05 & 50.70  \\
    NCE & 41.93 & 77.67 & 88.67 & 51.50  \\
    \bottomrule
    \end{tabular}
    \caption{Generalization evaluation results (\%).}
    \label{tb:union}
\end{table}

Besides MATH dataset, we evaluate the generalization ability of proposed framework. Following \citet{Specialize}, we only synthesize data and train the models on GSM8K and evaluate on all the four test sets. The higher performance of NAT and NCE on GSM8K indicates that the proposed method can generalize to previously commonly used dataset in the field of model specialization. NCE outperforms others in A-M-S datasets suggests that the calibrated dark knowledge from logits distributions can improve out-of-distribution (OOD) performance.

\subsubsection{Ablation study}
\begin{table}[t]
    \centering
    \small
        \begin{tabular}{l c c c c c c c}
    \toprule
    Methods & CP  & NT & PC  & G & A* &Ave\\ \midrule
    NAT & 5.70  &  6.67 & 3.85  & 5.64 & 7.72 & 6.81 \\
    - Neg Data & 6.55  & 4.63 & 4.40  & 3.97 & 7.88 & 6.58\\
    - Neg LoRA & 5.02  & 4.81 & 5.31  & 4.75 & 7.27 & 6.05\\
    - Att & 2.84  & 5.19 & 4.21  & 4.38 & 5.99 & 5.22\\ 
    - Dual & 7.38  & 5.37 & 4.21  & 4.80 & 6.87 & 6.30\\ 
 
    \bottomrule
    \end{tabular}
    \caption{Ablation study results (\%) for NAT.}
    \label{tb:ablation}
\end{table}

To demonstrate the necessity of each component in NAT, we take a series of ablation study by removing the following parts: (1) Neg Data: The whole dual LoRA structure and attention integration only on positive data. (2) Neg LoRA: Based on (1), the negative LoRA will be further removed. (3) Att: Instead of the attention mechanism, we integrate two LoRA modules by a gated function. (4) Dual: We modify the range of Equation~\ref{eq:attention1} to [0, 1] rather than [-0.5, 0,5], which means the knowledge from negative LoRA can only be absorbed from positive way.

The results are shown in Table~\ref{tb:ablation}. When filtering negative samples with same model structure, we find that model accuracy decreases, confirming the value of negative knowledge. Further removing the negative LoRA illustrates the importance of dual LoRA structure. The performance drops dramatically without attention mechanism, indicating it plays an important role in integrating LoRA modules. When changing the range of $\alpha$ to [0, 1], which forces positive LoRA to add the knowledge from negative LoRA without the minus option. The lower accuracy suggests that avoiding being influenced by undesirable behaviors while extracting useful knowledge from negative samples is necessary.

\subsubsection{Attention}
\begin{figure*}[t]
\centering
\includegraphics[width=0.98\textwidth]{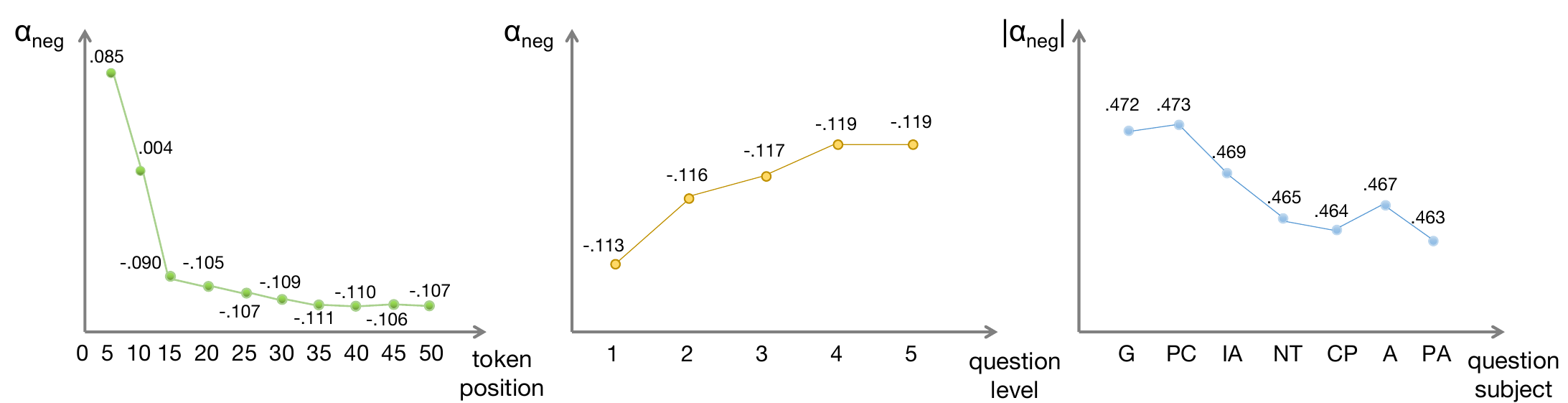} % Reduce the figure size so that it is slightly narrower than the column.
\caption{Analysis of $\alpha_{neg}$ along dimensions of: token position, question level, and question subject.
}
\label{pic:alpha}
\end{figure*}

To fully comprehend how knowledge from $\mathcal{M}_{neg}$ is integrated into $\mathcal{M}_{NAT}$ during the NAT process, we analyzed the averaged attention weights on the output of $\theta_{neg}$ ($\alpha_{neg}$ in Eq.~\eqref{eq:attention1}) along 3 dimensions: token position, question level, and question subject.

As shown in the Figure~\ref{pic:alpha}, as the position of the generated token increases, $\alpha_{neg}$ gradually decreases from positive values to negative values. This indicates that in the initial generation stage of the rationale, $\mathcal{M}_{neg}$ provides positive knowledge that helps establish the overall direction of problem solving. While in the fine-grained step generation stage, the undesirable behaviors provided by $\mathcal{M}_{neg}$ helps avoid similar errors. 
As for the question level, we observed that as the difficulty of the questions increases, the value of $\alpha_{neg}$ decreases. We hypothesize that this is because the undesirable behaviors in $\mathcal{M}_{neg}$ makes it challenging to accurately perform complex reasoning, leading it to primarily assist reasoning in difficult questions by providing negative references.
Finally, we sort the subjects according to the accuracy (ascending order) of $\mathcal{M}_{NAT}$ as the X-axis coordinates of the right subgraph of Figure~\ref{pic:alpha} to observe the tendency of $|\alpha_{neg}|$. Basically, $|\alpha_{neg}|$ gradually decreases with the difficulty of the subjects. Considering that the smaller $|\alpha_{neg}|$ is, the lesser the role $\mathcal{M}_{neg}$ plays in the NAT process ($|\alpha_{neg}|=0$ indicating that $\mathcal{M}_{neg}$ is not involved in the NAT phase), we believe that $\mathcal{M}_{neg}$ can play a greater role in addressing challenging subjects during NAT.

\subsubsection{Accuracy of Ranking Model}

\begin{figure}[t]
\centering
\includegraphics[width=0.4\textwidth]{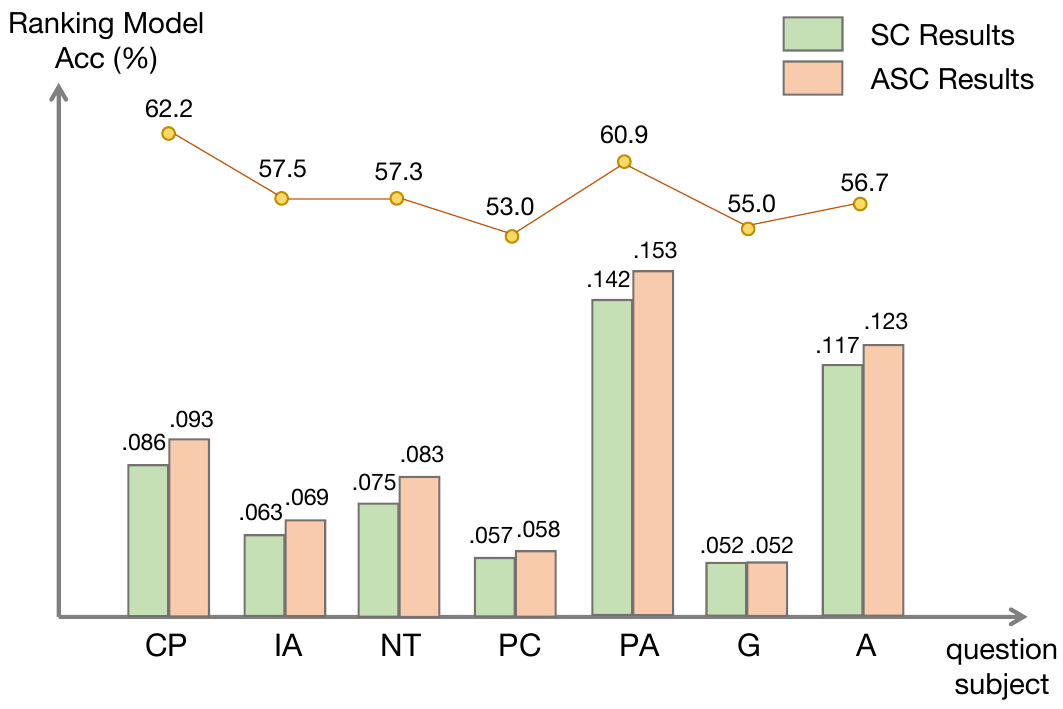} % Reduce the figure size so that it is slightly narrower than the column.
\caption{Relationship between the accuracy of distinguishing positive and negative rationales of $\mathcal{M}_{rank}$ and the improvement brought by ASC.
}
\label{pic:RM}
\end{figure}

We further explore the relationship between $\mathcal{M}_{rank}$'s accuracy of distinguishing positive and negative samples and the performance of ASC across 7 subjects. As shown in Figure~\ref{pic:RM}, we observed a positive correlation among the accuracy of $\mathcal{M}_{rank}$, SC results, and the improvement of ASC over SC. 
One potential reason is that relatively easier subjects can lead the reasoning model to achieve higher reasoning accuracy. This not only contributes to improved SC results but also provides more positive samples for training $\mathcal{M}_{rank}$. Through more comprehensive training, $\mathcal{M}_{rank}$ achieves higher accuracy, subsequently enhancing ASC's performance compared to SC. As current accuracy of $\mathcal{M}_{rank}$ is only around 60\% and yet it can significantly enhance the effectiveness of SC, we believe that there is substantial room for improvement in ASC.

\begin{figure}[t]
\centering
\includegraphics[width=0.45\textwidth]{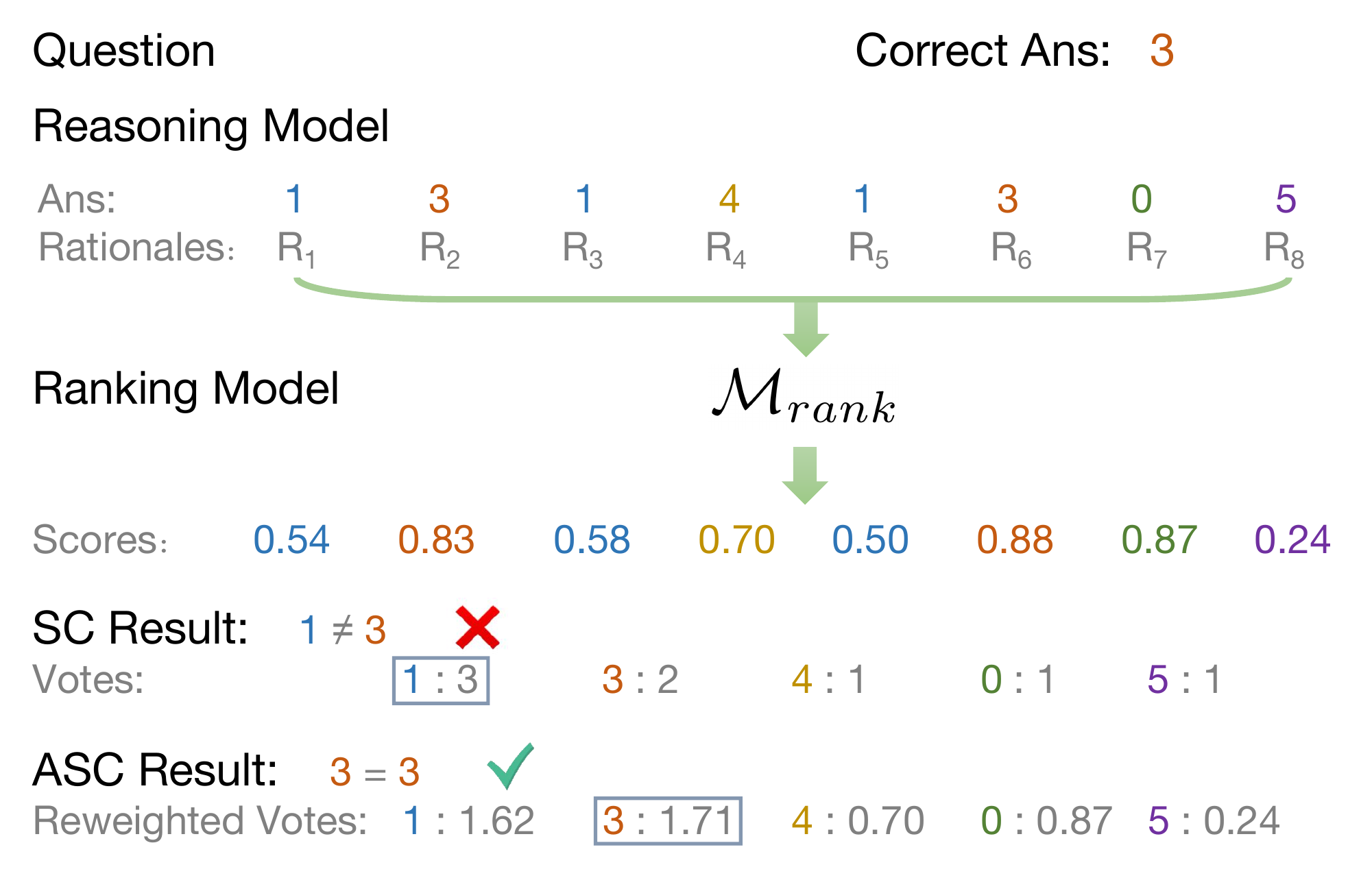} % Reduce the figure size so that it is slightly narrower than the column.
\caption{An intuitive example shows the strength of ASC.}
\label{pic:case}
\end{figure}

\subsubsection{Case Study about Adaptive Self-Consistency}
We provide an intuitive example (Figure~\ref{pic:case}) to show the superiority of ASC. In this example, deceptive candidate 1 is chosen as the prediction by SC due to having more votes than the correct candidate 3. $\mathcal{M}_{rank}$ adjusts the weights of the eight candidates based on rationales, resulting in the reweighted correct candidate 3 obtaining a higher vote count.

\section{Conclusion}
This work explores the effectiveness of negative data for distilling the complex reasoning ability from large language models to specialized small ones. 
We propose a novel framework, consisting of three progressive steps and fully leveraging the negative information through the entire process of model specialization. 
Negative assistant training can provide a more comprehensive way to employ the negative information from two aspects. 
Negative calibrated enhancement is able to calibrate the process of distillation, making it more targeted on crucial knowledge. 
Ranking model trained on two views of rationales can assign appropriate weights for answer aggregation to achieve adaptive self-consistency. 
Extensive experiments demonstrate that our framework can improve the effectiveness of distilling reasoning ability by the generated negative samples. 
\section{Acknowledgments}
This work is supported by Beijing Natural Science Foundation (No.4222037, L181010).

% \clearpage
\bibliography{aaai24}

% \clearpage
\appendix
\section{Appendix}
\subsection{Dataset}
\begin{table}[th]
    \centering
    \small
        \begin{tabular}{l c c c}
    \toprule
    Dataset & Subjects & Train & Test\\ \midrule
    \multirow{7}{*}{MATH}&Probability &  771  & 474  \\
    &InterAlgebra &  1295  & 903  \\ 
    &NumberTheory &  869  & 540  \\
    &Precalculus &  746  & 546 \\
    &Prealgebra &  1205  & 871 \\
    &Geometry &  870  & 479 \\
    &Algebra &  1744  & 1187  \\ \midrule
    GSM8K & - & 7473 & 1319\\
    ASDiv & - & - & 1218 \\
    MultiArith & - & - & 600\\
    SVAMP & - &  - & 1000\\
    \bottomrule
    \end{tabular}
    \caption{Statistical information about the datasets utilized in our experiments. we focus on MATH dataset and the four datasets below are for generalization evaluation.}
    \label{tb:data}
\end{table}

Table~\ref{tb:data} shows the detailed data statistics. In this study, we focus on MATH dataset and the four datasets below are intended for the purpose of evaluating generalization. For saving space, subjects in experiments section will be shown in short form. CP, IA, NT, PC, PA, G and A denote Counting and Probability, Intermediate Algebra, Number Theory, Precalculus, Prealgebra, Geometry and Algebra respectively. 

\subsection{Details of Baselines}
For learning from negative views, four baseline methods will be introduced as follows:
\begin{itemize}
\item \textbf{MIX} directly trains LLaMA with the mixture of both positive data $\mathcal{D}_{pos}$ and negative data $\mathcal{D}_{neg}$  by maximizing the following expectation:

\begin{equation}
\mathbb{E}_{(x,\hat{r},\hat{y}) \sim \mathcal{D}_{pos+neg}}  \mathrm{log}  P (\hat{y},\hat{r} | x;\theta).
\end{equation}
\item \textbf{CL} (contrastive learning) learns a representation of data such that the problem $x$ and rationales $r$ with answers $y$ of positive samples are close together in the representation space, while negative samples are far apart. In this work, the following expectation is maximized:

\begin{equation}
\small
\mathcal{L}_{CL}=-\mathrm{log}\frac{e^{sim(x,\hat{y}^{+}, \hat{r}^{+})}}{e^{sim(x,\hat{y}^{+}, \hat{r}^{+})}+\sum_{i}^{n}e^{sim(x,\hat{y}^{-}_i, \hat{r}^{-}_i)} }
\label{equation: CL}
\end{equation}

\begin{align}
\nonumber  \mathbb{E}_{(x,\hat{r}^{+},\hat{y}^{+}) \sim \mathcal{D}_{pos}}  \mathrm{log}  P (\hat{y},\hat{r} | x;\theta) \\
+ \mathbb{E}_{(x,\hat{r}^{-},\hat{y}^{-}) \sim \mathcal{D}_{neg},(x,\hat{r}^{+}, \hat{y}^{+})\sim\mathcal{D}_{pos}}\mathcal{L}_{CL},
\end{align}
% where we simply set $\lambda_1$ to 1.

\item \textbf{NT} (negative training) \citep{NT} conducts negative updates with training signals from negative samples to avoid model generating such data. The training objective is to maximize the following expectation:

\begin{align}
\nonumber  \mathbb{E}_{(x,\hat{r},\hat{y}) \sim \mathcal{D}_{pos}}  \mathrm{log}  P (\hat{y},\hat{r} | x;\theta) \\
 - \lambda_2 * \mathbb{E}_{(x,\hat{r},\hat{y}) \sim \mathcal{D}_{neg}}    \mathrm{log}  P  (\hat{y}, & \hat{r} | x;\theta).
\end{align}
The mixing hyper-parameter $\lambda_2$ is searched in $[0.05, 0.1, 1]$, and 0.05 is selected for its best performance.

\item \textbf{UL} (unlikelihood loss) \citep{UL} penalizes the model for outputting words with certain characteristics by introducing an unlikelihood loss term. In this work, we just penalize the negative samples in sentence level:

\begin{align}
\nonumber  \mathbb{E}_{(x,\hat{r},\hat{y}) \sim \mathcal{D}_{pos}}  \mathrm{log}  P (\hat{y},\hat{r} | x;\theta) \\
 + \lambda_3 * \mathbb{E}_{(x,\hat{r},\hat{y}) \sim \mathcal{D}_{neg}}    \mathrm{log}  P  (1 -\hat{y}, & \hat{r} | x;\theta). 
\end{align}
Similar to NT, we search the mixing hyper-parameter $\lambda_3$ in $[0.05, 0.1, 1]$, and 0.05 is selected as the best.

\end{itemize}

\subsection{Generalization Ability to Other Reasoning Tasks}
We selected MATH as our primary dataset for its difficulty and significance of transferring the capability from LLMs to small models.
In this paper, we have verified the generalization ability to other simpler datasets including GSM8K, ASDiv, MultiArith and SVAMP. Here we also conduct experiments on commonsense reasoning with StrategyQA dataset, the results from Table~\ref{tb:appendix-general} can demonstrate the generalization ability of NAT to other reasoning tasks.

\begin{table}[th]
\setlength{\abovecaptionskip}{0.3cm}
\renewcommand{\arraystretch}{1.2}
    \centering
    \small
        \begin{tabular}{l c c c c}
    \toprule
    Models & Data Source &Methods & Accuracy \\ \cline{1-4}
   \multirow{5}{*}{\makecell[c]{LLaMA 7B}} & Raw &Fine-tune& 61.2 +0\%\\\cline{2-4}
   &\multirow{2}{*}{\makecell[c]{GPT-3.5-Turbo}}&CoT KD & 69.2 +13.0\%\\
   &&NAT & 70.4 +15.0\%\\\cline{2-4}
   &\multirow{2}{*}{\makecell[c]{GPT-4}}&CoT KD & 72.4 +18.3\%\\
   &&NAT & 72.8 +19.0\%\\
    \bottomrule
    \end{tabular}
    \caption{Experimental results (\%) on StrategyQA.}
    \label{tb:appendix-general}
\end{table}

\subsection{NCE Analysis}

\begin{figure*}[t]
\centering
\includegraphics[width=0.98\textwidth]{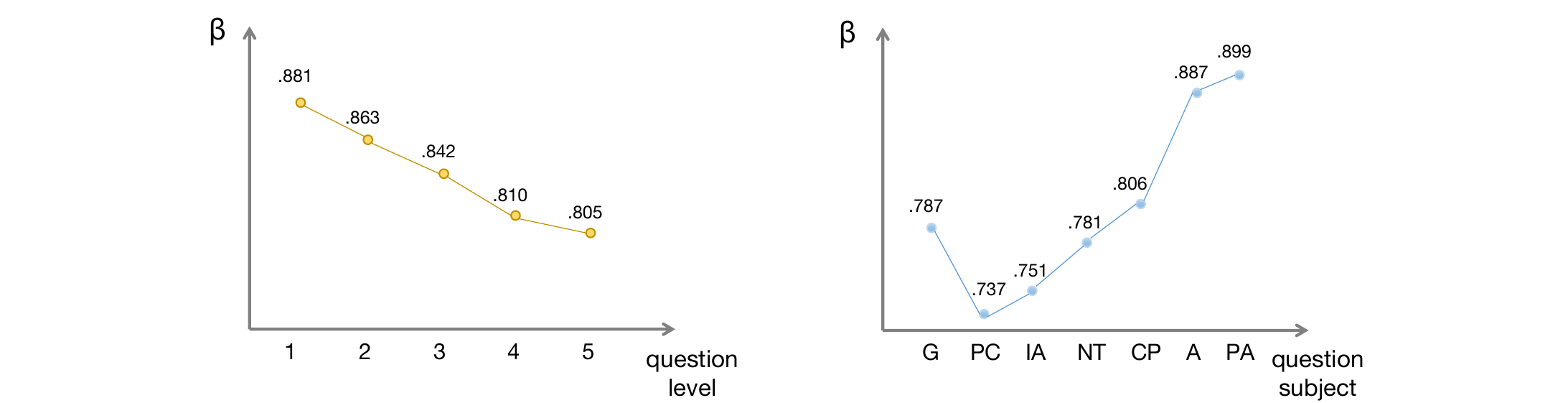} 
\caption{Analysis of $\beta$ along dimensions of: question level and question subject.
}
\label{pic:beta}
\end{figure*}

In order to further understand the effectiveness of NCE, we conducted experiments to analyze the relationship between $\beta$ and question level, as well as the subject to which the question belongs (Shown in Figure~\ref{pic:beta}). As for question level, we found that as the difficulty of the question increases, the $\beta$ gradually decreases, which means that the divergence between $\mathcal{M}_{neg}$ and $\mathcal{M}_{NAT}$ decreases accordingly. We conjecture that $\mathcal{M}_{neg}$ and $\mathcal{M}_{NAT}$ have a high degree of perplexity when solving difficult questions, so the probability distribution is relatively flat and the KL divergence is close. When answering simple questions, divergences arise due to their respective strengths (Table~\ref{tb:intro}). As for subjects, we found that in addition to Geometry, $\beta$ also basically increases as the difficulty of the subject decreases. The exception to Geometry indicates that difficulty is not the only factor determining $\beta$. Thus we do not have sufficient prior knowledge to determine $\beta$ and using KL divergence is necessary.

\begin{table}[th]
    \centering
    \small
    \begin{tabular}{l c c c c}
    \toprule
    &  Intersection & NAT & Neg &IoU\\ \midrule
    Probability &  6  & 26  &  16 & 0.167\\
    InterAlgebra &  6  & 44  &  21 & 0.102\\ 
    NumberTheory &  2  & 36  & 17  &0.039\\
    Precalculus &  3  & 21  &  16 &0.057\\
    Prealgebra &  15  & 83  & 43  & 0.135\\
    Geometry &  2  & 26  &  10 & 0.057\\
    Algebra &  16  & 92  &  43 & 0.134\\
    Overall &  49  & 328  &  166 & 0.110\\
    
        \bottomrule
    \end{tabular}
    \caption{The distribution of correct samples in MATH test set. Neg refers to model only trained by negative samples, and NAT refers to proposed method.}
    \label{tb:union2}
\end{table}

\begin{table*}[t]
\centering
\begin{tabular}{p{17cm}}

\toprule
\textbf{system:} Think the question step by step and give the answer. \\
\textbf{user:} Problem:
Find the domain of the expression $\frac{\sqrt{x-2}}{\sqrt{5-x}}$. \\
\textbf{assistant:} Solution:
The expressions inside each square root must be non-negative. Therefore, $x-2 \ge 0$, so $x\ge2$, and $5 - x \ge 0$, so $x \le 5$. Also, the denominator cannot be equal to zero, so $5-x>0$, which gives $x<5$. Therefore, the domain of the expression is $\boxed{[2,5)}$. \\
\textbf{user:} Problem:
Compute: $55\times1212-15\times1212$ . \\
\textbf{assistant:} Solution:
We have $55 \times 1212 - 15 \times 1212 = 1212(55-15) = 1212(40) = 4848(10) = \boxed{48480}$. \\
\textbf{user:} Problem:
Compute $\dbinom{16}{15}$. \\
\textbf{assistant:} Solution:
$\dbinom{16}{15}=\dbinom{16}{1}=\boxed{16}.$ \\
\textbf{user:} Problem:
Find $x$, where $x$ is a square number which is divisible by four, and between 39 and 80. \\
\textbf{assistant:} Solution:
We know that $x$ is between 39 and 80, and since $6^2 = 36 < 39$ and $9^2 = 81 > 80$, this means that $6^2 < x < 9^2$. This leaves us with two possibilities for $x$, which are $7^2 = 49$, and $8^2 = 64$. We then see that only 64 is divisible by four, so $x =$ $\boxed{64}$. \\
\textbf{user:} Problem:
Solve the inequality
\[\frac{(x - 2)(x - 3)(x - 4)}{(x - 1)(x - 5)(x - 6)} > 0.\] \\
\textbf{assistant:} Solution:
We can build a sign chart, but since all of the factors are linear, we can track what happens to the expression as $x$ increases.  At $x = 0,$ the expression is positive.  As $x$ increases past 1, the expression becomes negative.  As $x$ increases past 2, the expression becomes positive, and so on.  Thus, the solution is
\[x \in \boxed{(-\infty,1) \cup (2,3) \cup (4,5) \cup (6,\infty)}.\] \\
\textbf{user:} Problem:
A right circular cone has a volume of $12\pi$ cubic centimeters. The height of the cone is 4 cm. How many centimeters is the circumference of the base of the cone, in terms of $\pi$? \\
\textbf{assistant:} Solution:
The volume of a cone is $\frac{1}{3}\pi r^2 h$. We are given that the volume is $12\pi$ and the height is $4$. Thus, $\frac{1}{3}\pi r^2 \cdot 4 = 12\pi$. Solving for $r$, we find $r = 3$. Therefore, the circumference of the base is $2\pi r = \boxed{6\pi}$. \\
\textbf{user:} Problem:
How many perfect squares less than 1000 have a ones digit of 2, 3 or 4? \\
\textbf{assistant:} Solution:
Checking the squares from $1^2$ to $10^2$, we see that no squares end in 2 or 3, while a square ends in 4 if its square root ends in 2 or 8.  Since $31^2 < 1000 < 32^2$, we see that the squares less than 1000 ending in 4 are $2,8,12,18,22,28$.  Thus the desired answer is $\boxed{6}$. \\
\textbf{user:} Problem:
The diagonals of a rhombus are $10$ inches and $24$ inches. What is the perimeter of the rhombus, in inches? \\
\textbf{assistant:} Solution:
The diagonals are perpendicular bisectors of each other, so therefore the side length of the rhombus can be calculated as $\sqrt{5^2+12^2} = 13$. Therefore, the perimeter of the rhombus is $4 \times 13 = \boxed{52}$ inches. \\

\bottomrule
\end{tabular}
\caption{Demonstrations used to obtain responses from GPT-3.5 Turbo (GPT-4).}
\label{tb:demon}
\end{table*}

\subsection{Overlap Study}

Our initial discovery was that the IoU of the models trained on $\mathcal{D}_{neg}$ and $\mathcal{D}_{pos}$ was very small (0.074) for the correct predictions. However, by integrating the knowledge in $\mathcal{D}_{neg}$ and $\mathcal{D}_{pos}$, the IoU of $\mathcal{M}_{neg}$ and $\mathcal{M}_{NAT}$ has increased by 48.6\% (0.110). This indicates that our proposed method can effectively utilize the knowledge in $\mathcal{D}_{neg}$ and truly achieve turning dust into gold.

\subsection{Chain-of-thought Prompt}
We provide the prompt to obtain responses from GPT-3.5 Turbo (GPT-4) in Table~\ref{tb:demon}. We follow \citet{diver} to randomly sampled eight samples from different subjects and levels in the training set of MATH datasets to form this prompt.

\end{document}